\documentclass{article} 
\usepackage{nips13submit_e,times}
\usepackage{hyperref}
\usepackage{url}
\usepackage{float}
\usepackage{amsmath}
\usepackage{tikz}
\usetikzlibrary{decorations.pathreplacing}
\usetikzlibrary{decorations}

\title{Investigating the Role of Prior Disambiguation in Deep-learning Compositional Models of Meaning}

\author{
Jianpeng Cheng \\
University of Oxford \\
\texttt{\small jianpeng.cheng@cs.ox.ac.uk} \\
\And
Dimitri Kartsaklis \\
University of Oxford \\
\texttt{\small kartsak@cs.ox.ac.uk} \\
\And
Edward Grefenstette \\
Google DeepMind\\
\texttt{\small etg@google.com}
}

\nipsfinalcopy 

\begin{document}

\maketitle

\vspace{-0.3cm}
\begin{abstract}
This paper aims to explore the effect of prior disambiguation on neural network- based compositional models, with the hope that better semantic representations for text compounds can be produced. We disambiguate the input word vectors before they are fed into a compositional deep net. A series of evaluations shows the positive effect of prior disambiguation for such deep models.
\end{abstract}

\section{Introduction}
\label{sec:intro}
\vspace{-0.1cm}

While distributed representations of meaning began largely at the word level, the need for representations of the semantics of larger units of text, from phrases to entire documents, is evident. Early attempts at compositionality in distributed representations used fixed algebraic operations such as vector addition and component-wise multiplication \cite{mitchell2008vector} to obtain semantic representations of larger units of text from their constitutent representations. More recently, models represented relational words as tensors of various orders and tensor contraction was adopted as the mean of composition \cite{baroni2010nouns,coecke2010mathematical,GrefenstetteThesis2013}. Alongside these principally multi-linear methods of composition, we are observing an emergence of non-linear neural network-based compositional approaches, which derive a sentence vector by recursively applying neural networks to each pair of word vectors \cite{socher2010learning,socher2011semi,hermann-blunsom:2013:ACL2013}.

Although their levels of sophistication vary, all compositional methods for distributed semantics share the same problem: they take ambiguous word vectors as input, where each token is represented by a single vector regardless of the actual number of senses that the word has, and of the context in which it appears. To solve this problem, Reddy et al.~\cite{reddy2011dynamic} propose to disambiguate each word vector before composition for simple additive and multiplicative compositional models. This idea has now been successfully tested on a series of multi-linear compositional distributional models by Kartsaklis and colleagues \cite{kartsaklis2013prior,karts_conll2013,kartsaklis2014resolving}. In this paper we move one step further, by evaluating the effectiveness of a prior disambiguation step on neural compositional models. In Section~\ref{sec:neuralnet} we discuss the models evaluated in this paper, followed by a description of the disambiguation procedure of \cite{kartsaklis2013prior}, adapted to these models, in Section~\ref{sec:disambig}. Experimental evaluation of this procedure is presented and discussed in Sections~\ref{sec:experiments}--\ref{sec:discussion}. We conclude in Section~\ref{sec:conclusion} that prior disambiguation has a positive effect on neural models of composition.

\section{Neural networks for composing meaning}
\label{sec:neuralnet}
\vspace{-0.1cm}

Deep learning algorithms are capable of modelling complex relationships between inputs and outputs in NLP \cite{socher2011semi,socher2013recursive,hermann-blunsom:2013:ACL2013}. In this paper we are interested in capturing the meaning of a sentence by composing the vectorial representations of the words therein. In the most generic form of such a composition, a neural network is applied to each pair of words $w_1$ and $w_2$:

\begin{equation}
\overrightarrow{v} = f(\textbf{W}[\overrightarrow{w_1}:\overrightarrow{w_2}]+b)
\end{equation}\label{general}

where $[\overrightarrow{w_1}:\overrightarrow{w_2}]$ denotes the concatenation of the two vectors assigned to the words, \textbf{W} and $b$ are model parameters, and \emph{f} is a non-linear activation function. The compositional result, $\overrightarrow{v}$, is a vector representing the meaning of the bigram, and can be used again as an input to compute the representation of a larger text constituent in a recursive fashion. This process continues until all vectors of the words in a sentence have been merged into a single vectorial representation, which will serve as a semantic representation of that sentence or phrase. This class of models is known as \textit{recursive neural networks} (RecNNs) \cite{socher2010learning}.

In a variation of the above structure, each intermediate composition is performed via an auto-encoder, instead of a feed-forward network. A \textit{recursive auto-encoder} (RAE) \cite{socher2011semi,hermann-blunsom:2013:ACL2013} learns to reconstruct the input, encoded via a hidden layer, as faithfully as possible. The state of hidden layer of the RAE can be used as a compressed representation of the two original inputs. Since the optimization is based on the reconstruction error, a RAE is trained in an unsupervised fashion.

\section{Combining NNs with context-based word sense disambiguation}
\label{sec:disambig}
\vspace{-0.1cm}

The usual practice in deep learning models of meaning is to use as inputs ambiguous word representations, in which all possible meanings of a token are merged into a single vector. In this paper we evaluate a new methodology, in which each input word is associated with a set of vectors, each representing different meanings of the word in the training corpus. As input to the compositional network, we select the most probable meaning vector for each word given its context.

\begin{figure}[hb!]
\centering
\footnotesize
\begin{tikzpicture}[shorten >=1 pt,->,draw=black!50, node distance=\layersep, scale=0.80]
    \tikzset{
    ncbar angle/.initial=90,
    ncbar/.style={
        to path=(\tikztostart)
        -- ($(\tikztostart)!#1!\pgfkeysvalueof{/tikz/ncbar angle}:(\tikztotarget)$)
        -- ($(\tikztotarget)!($(\tikztostart)!#1!\pgfkeysvalueof{/tikz/ncbar angle}:(\tikztotarget)$)!\pgfkeysvalueof{/tikz/ncbar angle}:(\tikztostart)$)
        -- (\tikztotarget)
    },
    ncbar/.default=0.5cm,
}
    \tikzset{round left paren/.style={ncbar=0.5cm,out=120,in=-120}}
    \tikzset{round right paren/.style={ncbar=0.5cm,out=60,in=-60}}
    \tikzstyle{cell}=[rectangle, draw=black!50, rounded corners=1.5mm, minimum width=0.3cm,minimum height = 1.2cm]
    \tikzstyle{wsd}=[rectangle,  draw=black!50, rounded corners=1.5mm, minimum width=1cm,minimum height = 4.5cm]
    \tikzstyle{hidden_cell}=[rectangle,  draw=black!50, minimum width=0.8cm,minimum height = 0.8cm]
    \tikzstyle{dots}=[circle,fill=black!90,  minimum size=0.1cm,inner sep=0pt]
    \tikzstyle{annot} = [text width=4em, text centered]
    \tikzstyle{equation1} = [text width=13em, text centered]
    \tikzstyle{equation2} = [text width=15em, text centered]
    \node[cell] (I-3) at (0,-4) {};
    \node[cell] (I-2) at (0,-2) {};
    \node[cell] (I-1) at (0, 0) {};
    \node[cell] (I-31) at (1,-4) {};
    \node[cell] (I-21) at (1,-2) {};
    \node[cell] (I-11) at (1, 0) {};
    \node[cell] (I-32) at (1.5,-4) {};
    \node[cell] (I-22) at (1.5,-2) {};
    \node[cell] (I-12) at (1.5, 0) {};
    \node[cell] (I-33) at (2,-4) {};
    \node[cell] (I-23) at (2,-2) {};
    \node[cell] (I-13) at (2, 0) {};
    \node[dots] at (0,0) {};
    \node[dots] at (0,0.5) {};
    \node[dots] at (0,-0.5) {};
    \node[dots] at (1,0) {};
    \node[dots] at (1,0.5) {};
    \node[dots] at (1,-0.5) {};
    \node[dots] at (1.5,0) {};
    \node[dots] at (1.5,0.5) {};
    \node[dots] at (1.5,-0.5) {};
    \node[dots] at (2,0) {};
    \node[dots] at (2,0.5) {};
    \node[dots] at (2,-0.5) {};
    \node[dots] at (0,-2) {};
    \node[dots] at (0,-1.5) {};
    \node[dots] at (0,-2.5) {};
    \node[dots] at (1,-2) {};
    \node[dots] at (1,-1.5) {};
    \node[dots] at (1,-2.5) {};
    \node[dots] at (1.5,-2) {};
    \node[dots] at (1.5,-1.5) {};
    \node[dots] at (1.5,-2.5) {};
    \node[dots] at (2,-2) {};
    \node[dots] at (2,-1.5) {};
    \node[dots] at (2,-2.5) {};
    \node[dots] at (0,-4) {};
    \node[dots] at (0,-3.5) {};
    \node[dots] at (0,-4.5) {};
    \node[dots] at (1,-4) {};
    \node[dots] at (1,-3.5) {};
    \node[dots] at (1,-4.5) {};
    \node[dots] at (1.5,-4) {};
    \node[dots] at (1.5,-3.5) {};
    \node[dots] at (1.5,-4.5) {};
    \node[dots] at (2,-4) {};
    \node[dots] at (2,-3.5) {};
    \node[dots] at (2,-4.5) {};
    \draw[decorate,decoration={brace,amplitude=5},-] (0.8,-0.75) -- (0.8,0.75);
    \draw[decorate,decoration={brace,amplitude=5},-]  (0.8,-2.75) -- (0.8,-1.25);
    \draw[decorate,decoration={brace,amplitude=5},-]  (0.8,-4.75) -- (0.8,-3.25);
    \draw [decorate,decoration={brace,mirror,amplitude=5},-] (2.2,-0.75) -- (2.2,0.75);
    \draw [decorate,decoration={brace,mirror,amplitude=5},-] (2.2,-2.75) -- (2.2,-1.25);
    \draw [decorate,decoration={brace,mirror,amplitude=5},-] (2.2,-4.75) -- (2.2,-3.25);
    \node[] at (0.5, -0.75) {,};
    \node[] at (0.5, -2.75) {,};
    \node[] at (0.5, -4.75) {,};
    \draw [black, thick,-] (-0.3, -0.75) to [round left paren ] (-0.3,0.75);
    \draw [black, thick,-] (2.4,-0.75) to [round right paren] (2.4,0.75);
    \draw [black, thick,-] (-0.3, -2.75) to [round left paren ] (-0.3,-1.25);
    \draw [black, thick,-] (2.4,-2.75) to [round right paren] (2.4,-1.25);
    \draw [black, thick,-] (-0.3, -4.75) to [round left paren ] (-0.3,-3.25);
    \draw [black, thick,-] (2.4,-4.75) to [round right paren] (2.4,-3.25);
    \node [wsd] (w) at (4 ,-2) {WSD};
    \node[cell] (d-3) at (6,-4) {};
    \node[cell] (d-2) at (6,-2) {};
    \node[cell] (d-1) at (6, 0) {};
    \node[dots] at (6,0) {};
    \node[dots] at (6,0.5) {};
    \node[dots] at (6,-0.5) {};
    \node[dots] at (6,-2) {};
    \node[dots] at (6,-1.5) {};
    \node[dots] at (6,-2.5) {};
    \node[dots] at (6,-4) {};
    \node[dots] at (6,-3.5) {};
    \node[dots] at (6,-4.5) {};
    \node[hidden_cell] (h-1) at (7.5,-3){};
    \node[cell] (p-1) at (9,-3){};
    \node[hidden_cell] (h-2) at (10.5,-1.5){};
    \node[cell] (p-2) at (12,-1.5){};
    \node[dots] at (9,-3) {};
    \node[dots] at (9,-2.5) {};
    \node[dots] at (9,-3.5) {};
    \node[dots] at (12,-1.5) {};
    \node[dots] at (12,-1) {};
    \node[dots] at (12,-2) {};
    \path (d-2) edge (h-1);
    \path (d-3) edge (h-1);
    \path (h-1) edge (p-1);
    \path (p-1) edge (h-2);
    \path (d-1) edge (h-2);
    \path (h-2) edge (p-2);
    \path (2.6,0) edge (3.4,0);
    \path (2.6,-2) edge (3.4,-2);
    \path (2.6,-4) edge (3.4,-4);
    \path (4.65,0) edge (5.82,0);
    \path (4.65,-2) edge (5.82,-2);
    \path (4.65,-4) edge (5.82,-4);
   \node[annot,left of=I-3, node distance=1cm]{$word_3$:};
   \node[annot,left of=I-2, node distance=1cm]{$word_2$:};
   \node[annot,left of=I-1, node distance=1cm]{$word_1$:};
   \node[annot,right of=p-2, node distance=1cm]{$sentence$};
   \node[annot, above of=I-1, node distance = 1cm]{$\overrightarrow{w}$};
   \node[annot, above of=I-11, node distance = 1cm]{$\overrightarrow{s_1}$};
   \node[annot, above of=I-12, node distance = 1cm]{$\overrightarrow{s_2}$};
   \node[annot, above of=I-13, node distance = 1cm]{$\overrightarrow{s_3}$};

\end{tikzpicture}
\caption{From ambiguous word vectors to an unambiguous sentence vector.}
\label{fig:main}
\end{figure}
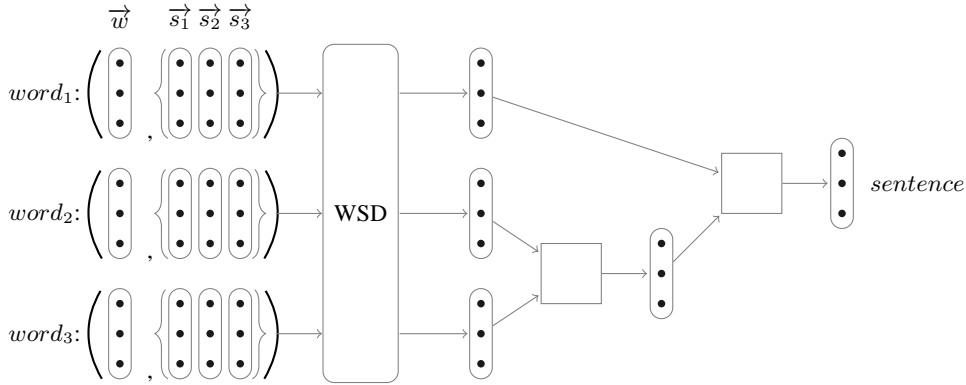

Our general methodology essentially recasts the approach of \cite{kartsaklis2013prior} in a deep learning setting; this is depicted in Figure~\ref{fig:main}. We first use a word sense induction step in order to discover the latent senses of each target word. For every occurrence of a target word $w_t$ in the corpus, we calculate a \textit{context vector} as the average of its neighbours, that is, $\overrightarrow{c_t} = \frac{1}{n}(\overrightarrow{w_1}+\overrightarrow{w_2}+\cdots+\overrightarrow{w_n})$
where $\overrightarrow{w_j}$ is the distributional (ambiguous) vector of the \emph{j}th neighbour.
After creating all the context vectors for the target word, we apply hierarchical agglomerative clustering to them in order to discover sensible groupings that hopefully correspond to different meanings of the word. As a vectorial representation for each meaning cluster, we use its centroid. Up to this point, each target word $w_t$ is associated to an ambiguous vector $\overrightarrow{w_t}$ and a set of 3 meaning vectors \emph{S}.



Assuming now an arbitrary word in some context $C$, we can select the most probable meaning vector for that word by creating a context vector $\overrightarrow{c_t}$ for $C$ as before (i.e. by averaging all the other words in $C$), and choosing the meaning vector which is the closest to $\overrightarrow{c_t}$.
For a set of meaning vectors $S$ and a distance metric $d(\overrightarrow{v},\overrightarrow{u})$, this is given as:

\begin{equation}
  \hat{\overrightarrow{v}} = \underset{\overrightarrow{v} \in S}{\arg\min}~d(\overrightarrow{v},\overrightarrow{c})
 \label{equ:wsd}
\end{equation}

Other works that combine NNs with WSD, but not in a compositional setting as here, are \cite{huang2012improving,neelakantanefficient}.

\section{Experiments}
\label{sec:experiments}
\vspace{-0.1cm}

In order to test the effect of prior disambiguation on deep learning compositional models, we disambiguate the constituent words of simple sentences of the form \textit{subject-verb-object} and verb phrases \textit{verb-object} before composition in a number of tasks. Furthermore, we evaluate two disambiguation strategies: in the first we disambiguate every word in a sentence, while in the second disambiguation applies only to verbs, which are usually the most ambiguous part of language. We evaluate the quality of the compositional results by measuring the similarity between sentence vectors---a good compositional model should be able to construct sentence vectors that reflect the true semantic relationships among the sentences.

Towards this purpose we use three phrase similarity datasets from the work of Grefenstette and Sadrzadeh \cite{grefenstette2011experimental} (G\&S), Kartsaklis et al.~\cite{karts_conll2013} (K\&S) and Mitchell and Lapata \cite{mitchell2010composition} (M\&L), consisting of pairs of sentences or phrases annotated with similarity scores by human evaluators. Our task is to measure to what extent the similarity computed by the composite vectors matches that of human judgements. In the first two datasets, which are based on \textit{subject-verb-object} structures, each pair of sentences is constructed around ambiguous verbs, while subject and object nouns are the same for the two sentences. The two datasets differ in the way ambiguous verbs were selected (in the former the selection was done automatically, while in the latter by humans), and in the fact that in the K\&S dataset every noun is modified by an appropriate adjective. For the M\&L dataset (comprised of \textit{verb-object} constructs) word ambiguity does not play a specific role, so from this aspect this dataset constitutes a more natural evaluation test for our models ``in the wild''.

In terms of neural composition models, we implement a RecNN and a RAE. Furthermore, we use simple additive and multiplicative models as baselines, where the representation of a sentence is derived by summing up the word vectors or taking the component-wise multiplication of them.

For each dataset and each model, the evaluation is conducted in two ways. First, we measure the Spearman's correlation between the computed cosine similarities of the composite sentence vectors and the corresponding human scores. Second, we apply a more relaxed evaluation, based on a binary classification task. Specifically, we use the human score that corresponds to each pair of sentences in order to decide a label for that pair (1 if the two sentences are highly similar and 0 otherwise), and we use the training set that results in from this procedure as input to a logistics regression classifier. We report the 4-fold cross validation accuracy as a measure of the matching rate. The results for each dataset and experiment are listed in the Tables~\ref{gs1}--\ref{ml2}.

\begin{table}[ht!]
  \centering
    \footnotesize
\begin{tabular}{|c|c|c|c|}
  \hline
  \multicolumn{4}{|c|}{Spearman's correlation} \\
  \hline\hline
    & No disambiguation & Disambig. every word & Disamb. verbs only \\
  \hline
  Additive model & \textbf{0.221} & 0.071 & 0.105 \\
  Multiplicative model & 0.085 & 0.012 & 0.043 \\
  RecNN & 0.127 & 0.119 & 0.128 \\
  RAE & 0.124 & 0.098 & 0.126 \\
  \hline\hline
  \multicolumn{4}{|c|}{Cross validation accuracy} \\
  \hline\hline
  Additive model & 63.07\% & 63.08\% & 62.48\% \\
  Multiplicative model & 61.89\% & 59.20\% & 60.11\% \\
  RecNN & 62.66\% & 63.53\% & \textbf{66.19\%} \\
  RAE & 63.04\% & 60.51\% & 65.17\% \\
  \hline
\end{tabular}
  \caption{Results for G\&S dataset.}\label{gs1}
\end{table}

\begin{table}[ht!]
  \centering
    \footnotesize
\begin{tabular}{|c|c|c|c|}
  \hline
  \multicolumn{4}{|c|}{Spearman's correlation} \\
  \hline\hline
    & No disambiguation & Disambig. every word & Disamb. verbs only \\
  \hline
  Additive model & 0.132 & \textbf{0.152} & 0.147 \\
  Multiplicative model & 0.049 & 0.129 & 0.104 \\
  RecNN & 0.085 & 0.098 & 0.101 \\
  RAE & 0.106 & 0.112 & 0.123 \\
  \hline\hline
  \multicolumn{4}{|c|}{Cross validation accuracy} \\
  \hline\hline
  Additive model & 49.28\% & 51.51\% & 51.04\% \\
  Multiplicative model & 49.76\% & 52.37\% & 53.06\% \\
  RecNN & 51.37\% & 52.64\% & \textbf{59.26\%} \\
  RAE & 50.92\% & 53.35\% & 59.17\% \\
  \hline
\end{tabular}
  \caption{Results for K\&S dataset.}\label{ka2}
\end{table}

\begin{table}[ht!]
  \centering
    \footnotesize
\begin{tabular}{|c|c|c|c|}
  \hline
  \multicolumn{4}{|c|}{Spearman's correlation} \\
  \hline\hline
    & No disambiguation & Disamb. every word & Disamb. verbs only \\
  Additive model & 0.379 & \textbf{0.407} & 0.382 \\
  Multiplicative model & 0.301 & 0.305 & 0.285 \\
  RecNN & 0.297 & 0.309 & 0.311 \\
  RAE & 0.282 & 0.301 & 0.303 \\
  \hline\hline
  \multicolumn{4}{|c|}{Cross validation accuracy} \\
  \hline\hline
  Additive model & 56.88\% & 59.31\% & 58.26\% \\
  Multiplicative model & 59.53\% & 59.24\% & 57.94\% \\
  RecNN & 60.17\% & 61.11\% & \textbf{61.20\%} \\
  RAE & 59.28\% & 59.16\% & 60.95\% \\
  \hline
\end{tabular}
  \caption{Results for M\&L dataset.}\label{ml2}
\end{table}

\section{Discussion}
\label{sec:discussion}

The results are quite promising, since they suggest that disambiguation as an extra step prior to composition can bring at least marginal benefits to deep learning compositional models. Comparing the numbers we got from the three datasets, the effect of disambiguation is clearest for the M\&L dataset. In both evaluations we carried out, disambiguation has a positive effect for the subsequent composition. This is very encouraging, since the words in this dataset were not chosen to be ambiguous on purpose. In other words, the results imply that for a generic sentence prior disambiguation can act as a useful pre-processing step, which might improve the final outcome (if the sentence has ambiguous words) or not (if all words are unambiguous), but never decrease the performance.

The effect of disambiguation seems also to be quite clear for the K\&S dataset, whereas the result for the G\&S dataset, although positive, is less definite. We speculate that the reason behind this difference is the way each dataset was constructed: for example, the K\&S dataset contains verbs and alternatives meanings of them that correspond to distinct \textit{homonymous} cases, e.g. such as verb `file' with alternative meanings `register' and `smooth'. On the other hand, the G\&S dataset contains many \textit{polysemous} cases, where there exist very slight variations in the senses of verbs, such as between the verb `write' and the alternative meanings `spell' and `publish'.



In terms of the comparison between deep learning models and algebraic baselines, the results are not very clear. Despite the well-known benefits of the deep learning methods in natural language processing, this work suggests for one more time that simple component-wise compositional operators might constitute a hard-to-beat baseline for certain tasks: Although the two deep learning models in general returned superior results for the second evaluation task, they could not beat the additive approach in the Spearman's correlation measure. In fact, similar findings have been reported previously in the study of Blacoe and Lapata \cite{blacoe2012comparison} and Kartsaklis et al.~\cite{kartsaklis2013prior}. However, when trying to interpret the effectiveness of the two approaches, we need to consider a generic scenario in which sentences and phrases are not restricted to a fixed length and structure. Obviously, the advantage of deep learning models would be more significant when dealing with longer text segments.


\section{Conclusion}
\label{sec:conclusion}

The main contribution of this paper is that it suggests that explicitly dealing with the issue of disambiguation can be an effective way to improve the performance of deep learning compositional models of meaning. For our simple approach of adding a prior disambiguation step to word vectors, the benefits are small. A reasonable future direction, then, would be to incorporate an explicit disambiguation step within the architecture of the compositional model, that deals with ambiguity during the training process itself. The current work indicates that such an approach, which is much more aligned with the concept of deep learning, could result in drastic improvements in the performance of a compositional model.

{\small
\bibliographystyle{abbrv}
\bibliography{nips}}

\end{document}